\documentclass[runningheads]{llncs}
\usepackage{graphicx}
\usepackage{times}
\usepackage{latexsym}
\usepackage{booktabs}
\usepackage[nolist]{acronym}
\usepackage{url}
\usepackage{todonotes}
\usepackage{array}

\usepackage{tabularx}
\usepackage{dcolumn}
\usepackage{longtable}
\usepackage{supertabular}
\usepackage[inline]{enumitem}
\usepackage{xspace}
\usepackage{listings}
\usepackage{fancyvrb}
\usepackage{comment}
\usepackage{wrapfig}  
\usepackage{caption}
\usepackage{longtable}
\usepackage{multirow}
\usepackage{colortbl}
\usepackage[T1]{fontenc}
\usepackage{tikz}
\usepackage{footnote}
\usepackage{amssymb}
\usepackage[ruled,vlined,linesnumbered]{algorithm2e}
\usepackage{url}
\usepackage{mathtools}
\usepackage{verbatim}
\usepackage{lipsum}
\usepackage{listingsutf8}
\usepackage{lscape}
\usepackage{pifont}
\usepackage{numprint}
\usepackage{adjustbox}
\usepackage{multirow}
\usepackage{multicol}
\usepackage{float}
\usepackage{pifont}
\newcommand{\triple}[3]{$\langle$\texttt{#1},\ \texttt{#2},\ \texttt{#3}$\rangle$}

\usepackage{tikz,xcolor,hyperref}

\definecolor{lime}{HTML}{A6CE39}
\DeclareRobustCommand{\orcidicon}{
	\begin{tikzpicture}
	\draw[lime, fill=lime] (0,0) 
	circle [radius=0.16] 
	node[white] {{\fontfamily{qag}\selectfont \tiny ID}};
	\draw[white, fill=white] (-0.0625,0.095) 
	circle [radius=0.007];
	\end{tikzpicture}
	\hspace{-2mm}
}

\foreach \x in {A, ..., Z}{\expandafter\xdef\csname orcid\x\endcsname{\noexpand\href{https://orcid.org/\csname orcidauthor\x\endcsname}
			{\noexpand\orcidicon}}
}

\makeatletter

\renewcommand*{\@fnsymbol}[1]{\ensuremath{\ifcase#1\or *\or \dagger\or \ddagger\or
    \mathsection\or \mathparagraph\or \|\or **\or \dagger\dagger
    \or \ddagger\ddagger \else\@ctrerr\fi}}

\newcommand{\printfnsymbol}[1]{%
  \textsuperscript{\@fnsymbol{#1}}%
}
\makeatother

\begin{document}
\begin{acronym}[UML]
	\acro{AOS}{Agricultural Ontology Services}
	\acro{AGRIS}{Agricultural Science and Technology}
	\acro{API}{Application Programming Interface}
	\acro{AOS}{Agricultural Ontology Services}
	\acro{AGRIS}{Agricultural Science and Technology}
	\acro{API}{Application Programming Interface}
	\acro{A2KB}{Annotation to Knowledge Base}
	\acro{AI}{Artificial Intelligence}
	\acro{BPSO}{Binary Particle-Swarm Optimization}
	\acro{BPMLOD}{Best Practices for Multilingual Linked Open Data}
	\acro{BPSO}{Binary Particle-Swarm Optimization}
	\acro{BPMLOD}{Best Practices for Multilingual Linked Open Data}
	\acro{BFS}{Breadth-First-Search}
	\acro{BPE}{Byte Pair Encoding}
	\acro{BoW}{Bag-of-Words}
	\acro{CBD}{Concise Bounded Description}
	\acro{COG}{Content Oriented Guidelines}
	\acro{CSV}{Comma-Separated Values}
	\acro{CBMT}{Corpus-Based Machine Translation}
	\acro{CLIR}{Cross-Language Information Retrieval}
	\acro{CNN}{Convolutional Neural Network}
	\acro{DPSO}{Deterministic Particle-Swarm Optimization}
	\acro{DALY}{Disability Adjusted Life Year}
	\acro{DBMS}{Relational Database Management System}

	\acro{ER}{Entity Resolution}
	\acro{EM}{Expectation Maximization}
	\acro{EBMT}{Example-Based Machine Translation}
	\acro{EBNF}{Extended Backus--Naur Form}
	\acro{EL}{Entity Linking}
	\acro{FAO}{Food and Agriculture Organization of the United Nations}
	\acro{GIS}{Geographic Information Systems}
	\acro{GHO}{Global Health Observatory}
	\acro{GRU}{Gated recurrent unit}
	\acro{HDI}{Human Development Index}
	\acro{ICT}{Information and communication technologies}
	\acro{IFRS}{International Financial Reporting Standards}
	\acro{ICD}{International Classification of Diseases}
	\acro{IT}{Information Technology}
    \acro{KB}{Knowledge Base}
    \acro{KG}{Knowledge Graph}
    \acrodefplural{KG}{Knowledge Graphs}
    \acro{KGE}{Knowledge Graph Embeddings}
    \acro{KBSE}{Knowledge Base Semantic Embedding}
	\acro{LR}  {Language Resource}
	\acro{LD}  {Linked Data}
	\acro{LLOD}  {Linguistic Linked Open Data}
	\acro{LIMES}{LInk discovery framework for MEtric Spaces}
	\acro{LS}  {Link Specifications}
	\acro{LDIF}{Linked Data Integration Framework}
	\acro{LGD} {LinkedGeoData}
	\acro{LOD} {Linked Open Data}
	\acro{LOV} {Linked Open Vocabularies}
	\acro{LSTM}{Long Short-Term Memories}
	\acro{MSE}{Mean Squared Error}
	\acro{MWE}{Multiword Expressions}
	\acro{MT}{Machine Translation}
	\acro{ML}{Machine Learning}
	\acro{MR}{Machine Reading}
	\acro{NIF}{Natural Language Processing Interchange Format}
	\acro{NIF4OGGD}{NLP Interchange Format for Open German Governmental Data}
	\acro{NLP}{Natural Language Processing}
	\acro{NER}{Named Entity Recognition}
	\acro{NMT}{Neural Machine Translation}
	\acro{NN}{Neural Network}
	\acro{NLG}{Natural Language Generation}
	\acro{NED}{Named Entity Disambiguation}
	\acro{NERD}{Named Entity Recognition and Disambiguation}
	\acro{NL}{Natural Language}
	\acro{NIF}{NLP Interchange Format}
	\acro{NIF4OGGD}{NLP Interchange Format for Open German Governmental Data}
	\acro{NLP}{Natural Language Processing}
	\acro{NER}{Named Entity Recognition}
	\acro{NEL}{Named Entity Linking}
	\acro{NE}{Named Entity}
	\acro{NN}{Neural Network}
	\acro{NLI}{Natural Language Inference}
	\acro{OSM}{OpenStreetMap}
	\acro{OWL}{Web Ontology Language}
	\acro{OOV}{out-of-vocabulary}
	\acro{PFM}{Pseudo-F-Measures}
	\acro{PSO}{Particle-Swarm Optimization}
	\acro{PBSMT}{Phrase-Based Statistical Machine Translation}
	
	\acro{QA}{Question Answering}
	\acro{RDF}{Resource Description Framework}
	\acro{RBMT}{Rule-Based Machine Translation}
	\acro{RNN}{Recurrent Neural Network}
	\acro{ReLU}{rectified linear unit}
	\acro{RDFS}{RDF Schema}
	\acro{SKOS}{Simple Knowledge Organization System}
	\acro{SPARQL}{SPARQL Protocol and RDF Query Language}
	\acro{SRL}{Statistical Relational Learning}
	\acro{SWT}{Semantic Web Technologies}
	\acro{SW}{Semantic Web}
	\acro{SMT}{Statistical Machine Translation}
	\acro{SWMT}{Semantic Web Machine Translation}
	\acro{SKOS}{Simple Knowledge Organization System}
	\acro{SPARQL}{SPARQL Protocol and RDF Query Language}
	\acro{SRL}{Statistical Relational Learning}
	\acro{SF}{surface forms}

    \acro{TBMT} {Transfer-Based Machine Translation}
	\acro{UML}{Unified Modeling Language}
	\acro{USL}{Ukrainian Sign Language}
	\acro{URI}{Uniform Resource Identifier}
	\acro{WHO}{World Health Organization}
	\acro{WKT}{Well-Known Text}
	\acro{W3C}{World Wide Web Consortium}
	\acro{WSD}{Word Sense Disambiguation}
	\acro{WMT}{Workshop on Machine Translation}
    \acro{XML}{Extensible Markup Language}
	\acro{YPLL}{Years of Potential Life Lost}

	\acro{AOS}{Agricultural Ontology Services}
	\acro{AGRIS}{Agricultural Science and Technology}
	\acro{API}{Application Programming Interface}
	\acro{AOS}{Agricultural Ontology Services}
	\acro{AGRIS}{Agricultural Science and Technology}
	\acro{API}{Application Programming Interface}
	\acro{A2KB}{Annotation to Knowledge Base}
	\acro{BPSO}{Binary Particle-Swarm Optimization}
	\acro{BPMLOD}{Best Practices for Multilingual Linked Open Data}
	\acro{BPSO}{Binary Particle-Swarm Optimization}
	\acro{BPMLOD}{Best Practices for Multilingual Linked Open Data}
	\acro{BFS}{Breadth-First-Search}
	\acro{BPE}{Byte Pair Encoding}
	\acro{BoW}{Bag-of-Words}
	\acro{CBD}{Concise Bounded Description}
	\acro{COG}{Content Oriented Guidelines}
	\acro{CSV}{Comma-Separated Values}
	\acro{CBMT}{Corpus-Based Machine Translation}
	\acro{CLIR}{Cross-Language Information Retrieval}
	\acro{DPSO}{Deterministic Particle-Swarm Optimization}
	\acro{DALY}{Disability Adjusted Life Year}

	\acro{ER}{Entity Resolution}
	\acro{EM}{Expectation Maximization}
	\acro{EBMT}{Example-Based Machine Translation}
	\acro{EBNF}{Extended Backus--Naur Form}
	\acro{EL}{Entity Linking}
	\acro{FAO}{Food and Agriculture Organization of the United Nations}
	\acro{GIS}{Geographic Information Systems}
	\acro{GHO}{Global Health Observatory}
	\acro{GCN}{Graph Convolutional Network}
	\acro{GRU}{Gated recurrent unit}
	\acro{GAT}{Graph Attention Network}
	\acro{HDI}{Human Development Index}
	\acro{ICT}{Information and communication technologies}
	\acro{IFRS}{International Financial Reporting Standards}
	\acro{ICD}{International Classification of Diseases}
	\acro{IT}{Information Technology}
	\acro{IRI}{International Resource Identifier}
    \acro{KB}{Knowledge Base}
    \acro{KG}{Knowledge Graph}
    \acro{KGE}{Knowledge Graph Embeddings}
    \acro{KBSE}{Knowledge Base Semantic Embedding}
	\acro{LR}  {Language Resource}
	\acro{LD}  {Linked Data}
	\acro{LLOD}  {Linguistic Linked Open Data}
	\acro{LIMES}{LInk discovery framework for MEtric Spaces}
	\acro{LS}  {Link Specifications}
	\acro{LDIF}{Linked Data Integration Framework}
	\acro{LGD} {LinkedGeoData}
	\acro{LOD} {Linked Open Data}
	\acro{LSTM}{Long Short-Term Memories}
	\acro{MSE}{Mean Squared Error}
	\acro{MWE}{Multiword Expressions}
	\acro{MT}{Machine Translation}
	\acro{ML}{Machine Learning}
	\acro{MR}{Machine Reading}
	\acro{MOS}{Manchester OWL Syntax}
	\acro{NIF}{Natural Language Processing Interchange Format}
	\acro{NIF4OGGD}{NLP Interchange Format for Open German Governmental Data}
	\acro{NLP}{Natural Language Processing}
	\acro{NER}{Named Entity Recognition}
	\acro{NMT}{Neural Machine Translation}
	\acro{NN}{Neural Network}
	\acro{NLG}{Natural Language Generation}
	\acro{NED}{Named Entity Disambiguation}
	\acro{NERD}{Named Entity Recognition and Disambiguation}
	\acro{NL}{Natural Language}
	\acro{NIF}{NLP Interchange Format}
	\acro{NIF4OGGD}{NLP Interchange Format for Open German Governmental Data}
	\acro{NLP}{Natural Language Processing}
	\acro{NER}{Named Entity Recognition}
	\acro{NEL}{Named Entity Linking}
	\acro{NE}{Named Entity}
	\acro{NN}{Neural Network}
	\acro{NLI}{Natural Language Inference}
	\acro{OSM}{OpenStreetMap}
	\acro{OWL}{Web Ontology Language}
	\acro{OOV}{out-of-vocabulary}
	\acro{PFM}{Pseudo-F-Measures}
	\acro{PSO}{Particle-Swarm Optimization}
	\acro{PBSMT}{Phrase-Based Statistical Machine Translation}
	
	\acro{QA}{Question Answering}
	\acro{RDF}{Resource Description Framework}
	\acro{RBMT}{Rule-Based Machine Translation}
	\acro{RNN}{Recurrent Neural Network}
	\acro{ReLU}{rectified linear unit}
	\acro{SKOS}{Simple Knowledge Organization System}
	\acro{SPARQL}{SPARQL Protocol and RDF Query Language}
	\acro{SRL}{Statistical Relational Learning}
	\acro{SWT}{Semantic Web Technologies}
	\acro{SW}{Semantic Web}
	\acro{SMT}{Statistical Machine Translation}
	\acro{SWMT}{Semantic Web Machine Translation}
	\acro{SKOS}{Simple Knowledge Organization System}
	\acro{SPARQL}{SPARQL Protocol and RDF Query Language}
	\acro{SRL}{Statistical Relational Learning}
	\acro{SF}{surface forms}
	\acro{SVM}{Support Vector Machines}

    \acro{TBMT} {Transfer-Based Machine Translation}
	\acro{UML}{Unified Modeling Language}
	\acro{USL}{Ukrainian Sign Language}
	\acro{URI}{Uniform Resource Identifier}
	\acro{WHO}{World Health Organization}
	\acro{WKT}{Well-Known Text}
	\acro{W3C}{World Wide Web Consortium}
	\acro{WSD}{Word Sense Disambiguation}
	\acro{WWW}{World Wide Web}
    \acro{XML}{Extensible Markup Language}
	\acro{YPLL}{Years of Potential Life Lost}

	\acro{AOS}{Agricultural Ontology Services}
	\acro{AGRIS}{Agricultural Science and Technology}
	\acro{API}{Application Programming Interface}
	\acro{BPSO}{Binary Particle-Swarm Optimization}
	\acro{BPMLOD}{Best Practices for Multilingual Linked Open Data}
	\acro{CBD}{Concise Bounded Description}
	\acro{COG}{Content Oriented Guidelines}
	\acro{CSV}{Comma-Separated Values}
	\acro{CBMT}{Corpus-Based Machine Translation}
	\acro{CLIR}{Cross-Language Information Retrieval}
	\acro{DPSO}{Deterministic Particle-Swarm Optimization}
	\acro{DALY}{Disability Adjusted Life Year}

	\acro{ER}{Entity Resolution}
	\acro{EM}{Expectation Maximization}
	\acro{EBMT}{Example-Based Machine Translation}
	\acro{EBNF}{Extended Backus--Naur Form}
	\acro{EL}{Entity Linking}
	\acro{FAO}{Food and Agriculture Organization of the United Nations}
	\acro{GIS}{Geographic Information Systems}
	\acro{GHO}{Global Health Observatory}
	\acro{HDI}{Human Development Index}
	\acro{ICT}{Information and communication technologies}
    \acro{KB}{Knowledge Base}
    \acro{KBSE}{Knowledge Base Semantic Embedding}
	\acro{LR}  {Language Resource}
	\acro{LD}  {Linked Data}
	\acro{LLOD}  {Linguistic Linked Open Data}
	\acro{LIMES}{LInk discovery framework for MEtric Spaces}
	\acro{LS}  {Link Specifications}
	\acro{LDIF}{Linked Data Integration Framework}
	\acro{LGD} {LinkedGeoData}
	\acro{LOD} {Linked Open Data}
	\acro{MSE}{Mean Squared Error}
	\acro{MWE}{Multiword Expressions}
	\acro{MT}{Machine Translation}
	\acro{ML}{Machine Learning}
	\acro{NIF}{Natural Language Processing Interchange Format}
	\acro{NIF4OGGD}{NLP Interchange Format for Open German Governmental Data}
	\acro{NLP}{Natural Language Processing}
	\acro{NER}{Named Entity Recognition}
	\acro{NMT}{Neural Machine Translation}
	\acro{NN}{Neural Network}
	\acro{NLG}{Natural Language Generation}
	\acro{NED}{Named Entity Disambiguation}
	\acro{NERD}{Named Entity Recognition and Disambiguation}
	\acro{NL}{Natural Language}
	\acro{OSM}{OpenStreetMap}
	\acro{OWL}{Web Ontology Language}
	\acro{OOV}{out-of-vocabulary}
	\acro{PFM}{Pseudo-F-Measures}
	\acro{PSO}{Particle-Swarm Optimization}
	\acro{QA}{Question Answering}
	\acro{RDF}{Resource Description Framework}
	\acro{RBMT}{Ru\-le-Ba\-sed Ma\-chi\-ne Trans\-la\-tion}
	\acro{REG}{Referring Expression Generation}
	\acro{SKOS}{Simple Knowledge Organization System}
	\acro{SPARQL}{SPARQL Protocol and RDF Query Language}
	\acro{SRL}{Statistical Relational Learning}
	\acro{SWT}{Semantic Web Technologies}
	\acro{SW}{Semantic Web}
	\acro{SMT}{Statistical Machine Translation}
	\acro{SWMT}{Semantic Web Machine Translation}

    \acro{TBMT} {Transfer-Based Machine Translation}
	\acro{UML}{Unified Modeling Language}
	\acro{USL}{Ukrainian Sign Language}
	\acro{URL}{Uniform Resource Locator}
	\acro{WHO}{World Health Organization}
	\acro{WKT}{Well-Known Text}
	\acro{W3C}{World Wide Web Consortium}
	\acro{WSD}{Word Sense Disambiguation}
    \acro{XML}{Extensible Markup Language}
	\acro{YPLL}{Years of Potential Life Lost}

	\acro{AOS}{Agricultural Ontology Services}
	\acro{AGRIS}{Agricultural Science and Technology}
	\acro{API}{Application Programming Interface}
	\acro{AOS}{Agricultural Ontology Services}
	\acro{AGRIS}{Agricultural Science and Technology}
	\acro{API}{Application Programming Interface}
	\acro{A2KB}{Annotation to Knowledge Base}
	\acro{BPSO}{Binary Particle-Swarm Optimization}
	\acro{BPMLOD}{Best Practices for Multilingual Linked Open Data}
	\acro{BPSO}{Binary Particle-Swarm Optimization}
	\acro{BPMLOD}{Best Practices for Multilingual Linked Open Data}
	\acro{BFS}{Breadth-First-Search}
	\acro{BPE}{Byte Pair Encoding}
	\acro{BoW}{Bag-of-Words}
	\acro{CBD}{Concise Bounded Description}
	\acro{COG}{Content Oriented Guidelines}
	\acro{CSV}{Comma-Separated Values}
	\acro{CBMT}{Corpus-Based Machine Translation}
	\acro{CLIR}{Cross-Language Information Retrieval}
	\acro{DPSO}{Deterministic Particle-Swarm Optimization}
	\acro{DALY}{Disability Adjusted Life Year}

	\acro{ER}{Entity Resolution}
	\acro{EM}{Expectation Maximization}
	\acro{EBMT}{Example-Based Machine Translation}
	\acro{EBNF}{Extended Backus--Naur Form}
	\acro{EL}{Entity Linking}
	\acro{FAO}{Food and Agriculture Organization of the United Nations}
	\acro{GIS}{Geographic Information Systems}
	\acro{GHO}{Global Health Observatory}
	\acro{GRU}{Gated recurrent unit}
	\acro{HDI}{Human Development Index}
	\acro{ICT}{Information and communication technologies}
	\acro{IFRS}{International Financial Reporting Standards}
	\acro{ICD}{International Classification of Diseases}
	\acro{IT}{Information Technology}
    \acro{KB}{Knowledge Base}
    \acro{KG}{Knowledge Graph}
    \acro{KGE}{Knowledge Graph Embeddings}
    \acro{KBSE}{Knowledge Base Semantic Embedding}
	\acro{LR}  {Language Resource}
	\acro{LD}  {Linked Data}
	\acro{LLOD}  {Linguistic Linked Open Data}
	\acro{LIMES}{LInk discovery framework for MEtric Spaces}
	\acro{LS}  {Link Specifications}
	\acro{LDIF}{Linked Data Integration Framework}
	\acro{LGD} {LinkedGeoData}
	\acro{LOD} {Linked Open Data}
	\acro{LSTM}{Long Short-Term Memories}
	\acro{MSE}{Mean Squared Error}
	\acro{MWE}{Multiword Expressions}
	\acro{MT}{Machine Translation}
	\acro{ML}{Machine Learning}
	\acro{MR}{Machine Reading}
	\acro{NIF}{Natural Language Processing Interchange Format}
	\acro{NIF4OGGD}{NLP Interchange Format for Open German Governmental Data}
	\acro{NLP}{Natural Language Processing}
	\acro{NER}{Named Entity Recognition}
	\acro{NMT}{Neural Machine Translation}
	\acro{NN}{Neural Network}
	\acro{NLG}{Natural Language Generation}
	\acro{NED}{Named Entity Disambiguation}
	\acro{NERD}{Named Entity Recognition and Disambiguation}
	\acro{NL}{Natural Language}
	\acro{NIF}{NLP Interchange Format}
	\acro{NIF4OGGD}{NLP Interchange Format for Open German Governmental Data}
	\acro{NLP}{Natural Language Processing}
	\acro{NER}{Named Entity Recognition}
	\acro{NEL}{Named Entity Linking}
	\acro{NE}{Named Entity}
	\acro{NN}{Neural Network}
	\acro{NLI}{Natural Language Inference}
	\acro{OSM}{OpenStreetMap}
	\acro{OWL}{Web Ontology Language}
	\acro{OOV}{out-of-vocabulary}
	\acro{PFM}{Pseudo-F-Measures}
	\acro{PSO}{Particle-Swarm Optimization}
	\acro{PBSMT}{Phrase-Based Statistical Machine Translation}
	
	\acro{QA}{Question Answering}
	\acro{RDF}{Resource Description Framework}
	\acro{RBMT}{Rule-Based Machine Translation}
	\acro{RNN}{Recurrent Neural Network}
	\acro{ReLU}{rectified linear unit}
	\acro{SKOS}{Simple Knowledge Organization System}
	\acro{SPARQL}{SPARQL Protocol and RDF Query Language}
	\acro{SRL}{Statistical Relational Learning}
	\acro{SWT}{Semantic Web Technologies}
	\acro{SW}{Semantic Web}
	\acro{SMT}{Statistical Machine Translation}
	\acro{SWMT}{Semantic Web Machine Translation}
	\acro{SKOS}{Simple Knowledge Organization System}
	\acro{SPARQL}{SPARQL Protocol and RDF Query Language}
	\acro{SRL}{Statistical Relational Learning}
	\acro{SF}{surface forms}
	\acro{SVM}{Support Vector Machines}

    \acro{TBMT} {Transfer-Based Machine Translation}
	\acro{UML}{Unified Modeling Language}
	\acro{USL}{Ukrainian Sign Language}
	\acro{URI}{Uniform Resource Identifier}
	\acro{WHO}{World Health Organization}
	\acro{WKT}{Well-Known Text}
	\acro{W3C}{World Wide Web Consortium}
	\acro{WSD}{Word Sense Disambiguation}
    \acro{XML}{Extensible Markup Language}
	\acro{YPLL}{Years of Potential Life Lost}
\end{acronym}  
\title{NABU -- Multilingual Graph-based Neural RDF Verbalizer}
%
%
\author{Diego Moussallem~\orcidA{}\inst{1}\thanks{Equal contribution} \thanks{This work was carried out under the Google Summer of Code 2019} \and
Dwaraknath Gnaneshwar~\orcidB{}\inst{2}\printfnsymbol{1}\printfnsymbol{2}\and
Thiago Castro Ferreira~\orcidC{}\inst{3,4} \printfnsymbol{2} \and Axel-Cyrille Ngonga Ngomo~\orcidD{}\inst{1} }
\authorrunning{Moussallem et al.}
%
\institute{Data Science Group, University of Paderborn, Germany \\
\email{first.lastname@upb.de}\\ \and
DL group, Manipal Institute of Technology, India \\
\email{dwarakasharma@gmail.com}\\ \and
Federal University of Minas Gerais (UFMG), Brazil \\ \and
Tilburg center for Cognition and Communication (TiCC) \\
Tilburg University, The Netherlands \\
 \email{tcastrof@tilburguniversity.edu}
}

\maketitle              
\begin{abstract}
The RDF-to-text task has recently gained substantial attention due to continuous growth of Linked Data. In contrast to traditional pipeline models, recent studies have focused on neural models, which are now able to convert a set of RDF triples into text in an end-to-end style with promising results. However, English is the only language widely targeted. We address this research gap by presenting NABU, a multilingual graph-based neural model that verbalizes RDF data to German, Russian, and English. NABU is based on an encoder-decoder architecture, uses an encoder inspired by Graph Attention Networks and a Transformer as decoder. Our approach relies on the fact that knowledge graphs are language-agnostic and they hence can be used to generate multilingual text. We evaluate NABU in monolingual and multilingual settings on standard benchmarking WebNLG datasets. Our results show that NABU outperforms state-of-the-art approaches on English with 66.21 BLEU, and achieves consistent results across all languages on the multilingual scenario with 56.04 BLEU. 
\end{abstract}

\section{Introduction}

\ac{NLG} is the process of generating coherent natural language text from non-linguistic data~\cite{reiter2000building}. Despite community agreement on the text and speech output of these systems, there is far less consensus on what the input should be~\cite{gatt2017survey}. A large number of inputs have hence been employed for \ac{NLG} systems, including images \cite{xu2015show}, numeric data~\cite{gkatzia2014comparing}, and \ac{SW} data~\cite{Ngonga2019}. Practical applications can be found in domains such as weather forecasts \cite{mei2016}, feedback for car drivers \cite{braun2018}, diet management \cite{mazzei2018}.

Presently, the generation of natural language from  
\ac{RDF} data has gained substantial attention~\cite{bouayad2014natural}. The RDF-to-text task has hence been proposed to investigate the quality of automatically generated texts from \ac{RDF} \acp{KG}~\cite{colin2016webnlg}. 
With the emergence of neural methods, end-to-end data-to-text models have been introduced to learn input-output mappings directly. These approaches rely much less 
on explicit intermediate representations compared to rule-based approaches~\cite{gehrmann2018}. 

Although Neural \ac{NLG} models have been achieving very good results~\cite{gardent2017webnlg} 
, English is the only language that has been widely targeted.  
In this work, we alleviate this language limitation by proposing a multilingual approach, named NABU. The motivation behind multilingual models lies in several directions, mainly in (1) transfer learning; when low-resource language pairs are trained together with high-resource languages, the translation quality improves; (2) zero-shot translation, where multilingual models are able to translate between language pairs from similar families that were never seen during training; (3) Easy deploy, a multilingual model achieving same performance on many languages in comparison to several separate language-specific models are much more desirable for companies in terms of deployment~\cite{johnson2017google}.

Our approach, NABU, is based on the fact that knowledge graphs are language-agnostic and hence can be used on the encoder side to generate multilingual text. NABU consists of an encoder-decoder architecture which incorporates structural information of RDF triples using an encoding mechanism inspired by \ac{GAT}~\cite{velivckovic2017graph}. In contrast to recent related work~\cite{ribeiro2020modeling}, NABU relies on the use of a reification 
strategy for modeling the graph structure of RDF input. The decoder part 
is based on the vanilla Transformer model~\cite{vaswani2017} along with an unsupervised tokenization model. 


We evaluate NABU on the standard benchmarking WebNLG datasets\cite{gardent2017creating} in three settings: monolingual, bilingual and multilingual. For the monolingual setting, we compare NABU with state-of-the-art English approaches and also perform experiments on Russian and German. The goal of the bilingual setting is to analyze the performance of NABU for language families. To achieve this goal, we train and evaluate bilingual models using NABU on English-German and on English-Russian. In the multilingual setting, we compare NABU with a multilingual Transformer model on English, German and Russian.
Our results show that NABU outperforms state-of-the-art approaches on English and achieves 66.21 BLEU. NABU also achieves consistent results across all languages on multilingual settings with 56.04 BLEU. In addition, NABU presents promising results on the bilingual models with 61.99 BLEU. 
Our findings suggest that NABU is able to generate multilingual text with similar quality to that generated by humans.
The main contributions of this paper can be summarized as follows:

\begin{itemize}
    \item We present a novel approach dubbed NABU based on a \ac{GAT}-Transformer architecture for
generating multilingual text from RDF KGs.
    \item  NABU outperforms English state-of-the-art approaches with consistent average improvements of +10 BLEU, METEOR and chrF3 on the WebNLG datasets.
    \item NABU exploits the benefits of modeling of language families in the generation task.  
\end{itemize}

The version of NABU used in this paper and also all experimental data are publicly available.~\footnote{https://github.com/dice-group/NABU}.

\section{Related Work}

A significant body of research has investigated the generation of \ac{NL} texts from \ac{RDF} data. A plenty of research is based on template- and rule-based approaches such as~\cite{cimiano2013exploiting,duma2013generating,ell2014language,Ngonga2019}. Recently, the WebNLG \cite{colin2016webnlg} challenge made this research area more prominent by providing a benchmark corpus of English texts verbalizing RDF triples in 15 different semantic domains. Among the participating models, the works based on sequence-to-sequence \acp{NN} achieved some of the best results~\cite{sleimi2016generating,mrabet2016aligning}. Moreover, \ac{RDF} has also been showing promising benefits to the generation of benchmarks for evaluating \ac{NLG} systems~\cite{ngomo2018bengal}.

The choice of neural architectures for RDF-to-text has evolved constantly along the last couple of years. All end-to-end models submitted to the WebNLG challenge~\cite{gardent2017webnlg} received the set of triples in a linearized form as input. However, researchers have recently been experimenting with graph-based approaches, which take the RDF input formatted as a graph, with promising results. Marcheggiane and Perez~\cite{marcheggiani2018} proposed a structured data encoder based on \ac{GCN} that directly exploits the graph structure and presented better results than \ac{LSTM} models.
~Distiawan et al.~\cite{distiawan2018gtr} presented a GTR-LSTM architecture which captures the global information of a \ac{KG} by encoding the relationships both within a triple and between the triples. 
Ferreira et al.~\cite{ferreira2019neural} introduced a systematic comparison between neural pipeline and end-to-end data-to-text approaches for the generation of text from RDF triples. Although Marcheggiane and Perez~\cite{marcheggiani2018} showed that the linearisation of the input graph has several drawbacks, the authors implemented \ac{GRU} and Transformer architectures which showed results superior to those of the former architecture. 
Recently, Ribeiro et.al~\cite{ribeiro2020modeling} devised an unified graph attention network structure which investigates graph-to-text architectures that combined global and local graph representations to improve fluency in text generation. Their experiments demonstrated significant improvements on seen categories in the WebNLG dataset.

Despite the plethora of graph-based neural approaches on handling \ac{RDF} data, English is the only language which has been widely targeted. Recent efforts were made to create German and Russian language versions of WebNLG~\cite{ferreira2018enriching,shimorina2019creating}. However, no work that investigates these languages has been published at the time of writing. To the best of our knowledge, NABU is hence the first approach which tackles multilinguality in the RDF-to-text task.   

\section{The NABU Approach}

\begin{figure}[t]
\includegraphics[width=\textwidth]{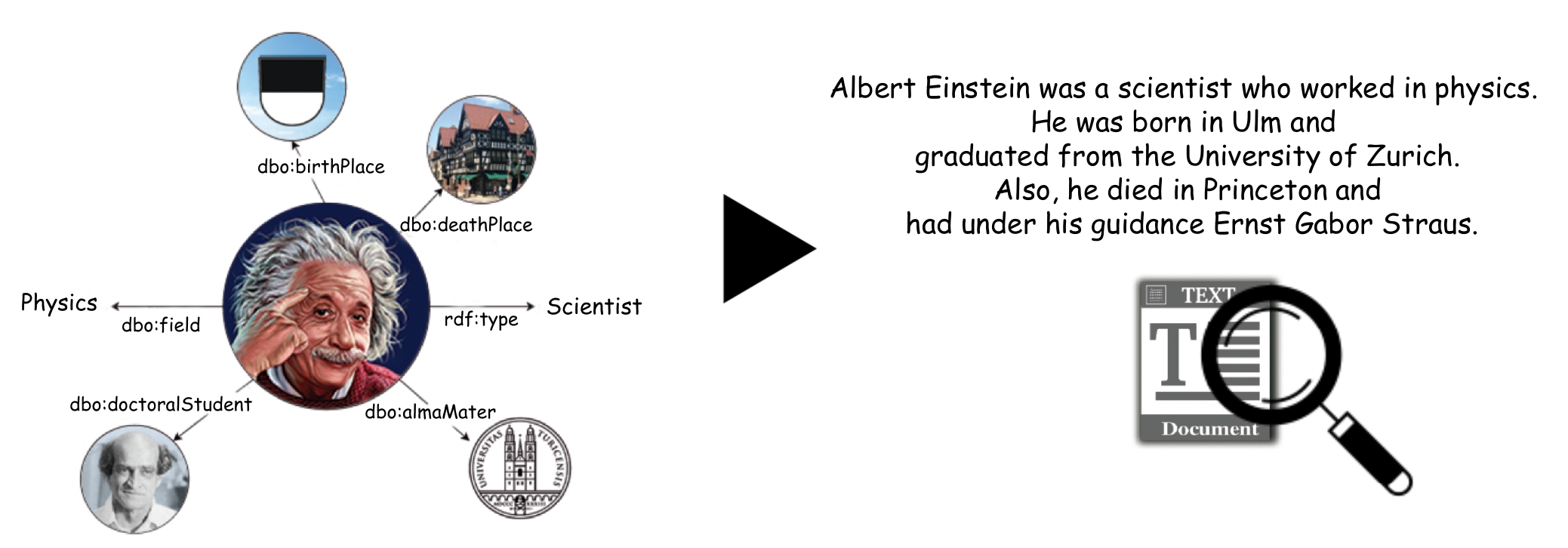}
\caption{Example of a set of triples (left) and the corresponding verbalization (right).}
\label{fig:RDF_example}
\end{figure}

NABU tackles RDF-to-text based on the formal description of a translation problem. The RDF-to-text task takes an RDF graph as input and generates an output text which reflects its 
meaning. Figure~\ref{fig:RDF_example} depicts an example of a set of 3 RDF triples and the corresponding text. Therefore, the underlying idea behind our approach is as follows: \emph{Given that \ac{KG}s are language-agnostic and represent facts often extracted from text, we can regard the facts (i.e., RDF triples) as sentences and train a model to translate the facts from a language-agnostic graph representation to several languages}. In the following, we give an overview of \ac{GAT} architecture and Transformer. Thereafter, we present NABU in detail. Throughout the description of our methodology and our experiments, we use DBpedia~\cite{auer2007dbpedia} as reference \ac{KB} since the benchmarking datasets are based on this \ac{KB}. 

\subsection{Background}

\subsubsection{Transformer}

Transformer-based models consist of an encoder and a decoder, i.e., a two-tier architecture where the encoder reads an input sequence $x=(x_1,...,x_n)$ and the decoder predicts a target sequence $y=(y_1,...,y_n)$. The encoder and decoder interact via a soft-attention mechanism~\cite{bahdanau2014neural,LuongAttention2015}, which comprises one or multiple attention layers. We follow the notations from Tang et al.~\cite{tang2018self} in the subsequent sections:
Let $m$ stand for the word embedding size and $n$ for the number of hidden units. Further, let $K$ be the vocabulary size of the source language. Then,
$h^{l}_{i}$ corresponds to the hidden state at step $i$ of layer $l$. $h^{l}_{i-1}$ represents the hidden state at the previous step of layer $l$ while $h^{l-1}_{i}$ means the hidden state at $i$ of layer $l-1$. $E \in \mathbb{R}^{m\times K}$ is a word embedding matrix, $W \in \mathbb{R}^{n\times m}$, $U \in \mathbb{R}^{n\times n}$ are weight matrices,  
$E_{x_{i}}$ refers to the embedding of $x_{i}$, and $e_{pos,i}$ indicates a 
positional embedding at position $i$. 

Transformer models rely deeply on self-attention networks. Each token is connected to every other token in the same sentence directly via self-attention. Thus, the path length between any two tokens is $1$. Due to lack of recurrence found in \ac{RNN}, Transformers implement 
\textit{positional encoding} to input and output. Additionally, these models rely on multi-head attention to feature attention networks, which are more complex in comparison to the $1$-head attention mechanism used in RNNs. In contrast to RNN, the positional information is also preserved in positional embeddings. Equation~\ref{eq:self-attention} describes the hidden state $h^{l}_{i}$, which is calculated from all hidden states of the previous layer. $f$ represents a feed-forward network with the \ac{ReLU} as the activation function and layer normalization. The first layer is implemented as $h^{0}_{i} = WE_{x_{i}} + e_{pos,i}$. Moreover, the decoder has a multi-head attention over the encoder's hidden states:

\begin{equation} \label{eq:self-attention}
h^{l}_{i} = h^{l-1}_{i} + f(\text{self-attention}(h^{l-1}_{i})).
\end{equation}

\subsubsection{Graph Attention Networks}

Deep Learning on non-euclidean data has recently gained substantial research interest due to the abundance of its availability. A plethora of problems can be solved efficiently by representing data in a data structure that can utilize the inherent structure and inter-entity relationships. Kipf and Welling~\cite{kipf2016semi} introduced \ac{GCN}, through which they generalize the convolution operation of \ac{CNN} to graph structures. Every layer in a \ac{GCN} has a weight matrix \textit{W} that transforms nodes feature vectors from a low-dimensional representation space to high-dimensional representation space, which aims to preserve the structure of the graph. 

Consider a graph of \textit{z} nodes 
and a set of node features ($\Vec{h_{1}}, \Vec{h_{2}},.., \Vec{h_{z}}$). A \ac{GCN} layer computes a net set of features ($\Vec{h_{1}^{'}}, \Vec{h_{2}^{'}},.., \Vec{h_{z}^{'}}$). First the feature matrix is multiplied with \textit{W} $\Vec{g} = W\Vec{h}$. Then, the aggregated sum of node features are normalized using normalization constant $ \frac{1}{c_{ij}}$ to stabilize the update rule. Finally, 
$$\Vec{h_{i}^{'}} = \sigma \left(\sum_{j \epsilon N_{i}} \frac{1}{c_{ij}} \Vec{g}_{j} \right)$$ 

However, the convolution operation in \ac{GCN} does not take into account the fact that some nodes are more important than others to generate a particular segment of the target sentence. To alleviate this problem, Velickovic et al.~\cite{velivckovic2017graph} devised \ac{GAT}, which converts the normalization constant into dynamic attention coefficients. The attention coefficients are calculated by applying \textit{self-attention} over node features. In one forward pass, a \ac{GAT} layer calculate a score of a given node that quantifies the importance of neighbors to its representation:
$$e_{ij} = a \left( \Vec{h_{i}}, \Vec{h_{j}} \right).$$
 
The attention scores are then normalized using softmax: 
$$\alpha_{ij} = \frac{exp(e_{ij})}{\sum_{k \epsilon N_{i}}exp(e_{ik})}.$$

\subsection{Approach}

\begin{figure}[tbh]
\includegraphics[width=\textwidth]{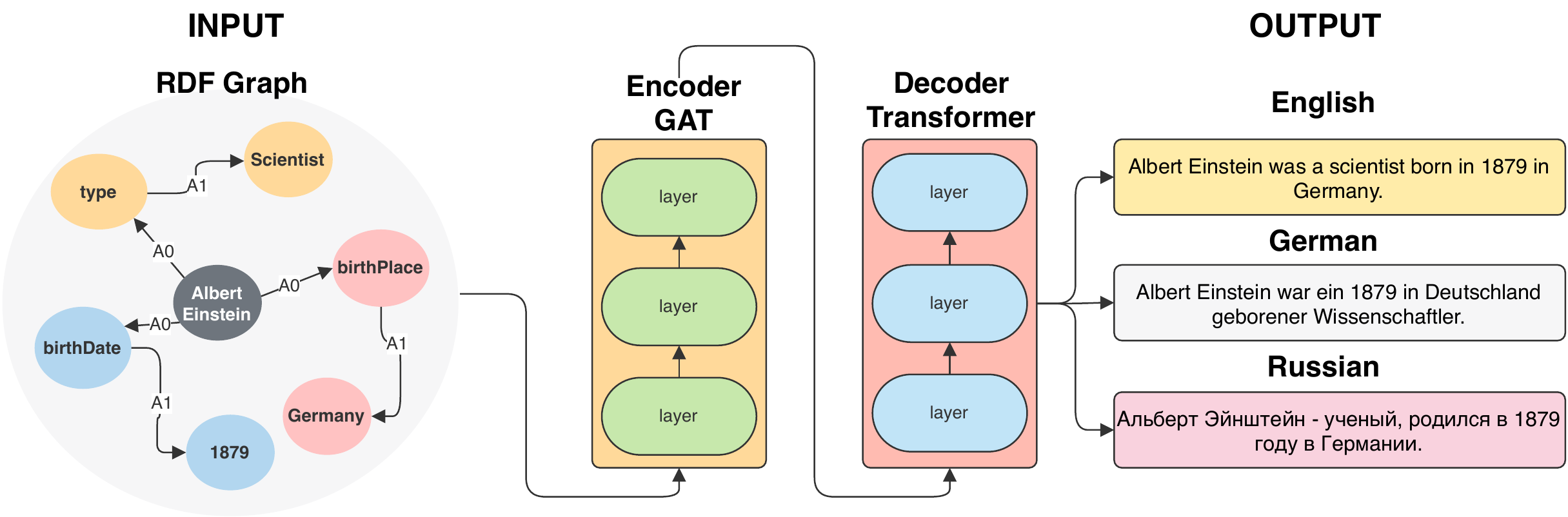}
\caption{NABU architecture}
\label{fig:NABU_arch}
\end{figure}

Graph-based \ac{NN}s have been used successfully to parse and support the generation of natural-language sentences from RDF \ac{KG}. Although \ac{GAT} models have shown to alleviate the loss of node information, the network still suffers from parameter explosion depending on the size of the graph structure~\cite{beck2018graph}. To alleviate the parameters explosion problem, 
we follow the same strategy used in \cite{marcheggiani2018}, named reification,\footnote{Not to be confused with RDFS reification.}  to slightly modify how the RDF graph is encoded. We describe below how reification is applied. Afterward, we explain the encoder and decoder parts of NABU. An overview of NABU architecture after reification can be found in Figure \ref{fig:NABU_arch}.

\subsubsection{Reification.} RDF triples are represented as a graph in which (i) the subjects and objects are nodes and (ii) predicates (relationships) between them are labeled edges. For example, \texttt{<Albert\-\_Einstein, birthPlace, Germany>} can be seen as a sub-\ac{KG} in DBpedia where \texttt{Albert\_ Einstein} and \texttt{Germany} are the nodes and \texttt{birthPlace} is the edge. However, the edges are encoded as parameters by the \ac{GAT}, and the parameters explosion problem stated by Beck et al.~\cite{beck2018graph} often occurs. 

\begin{figure}[tbh]
\includegraphics[width=\textwidth]{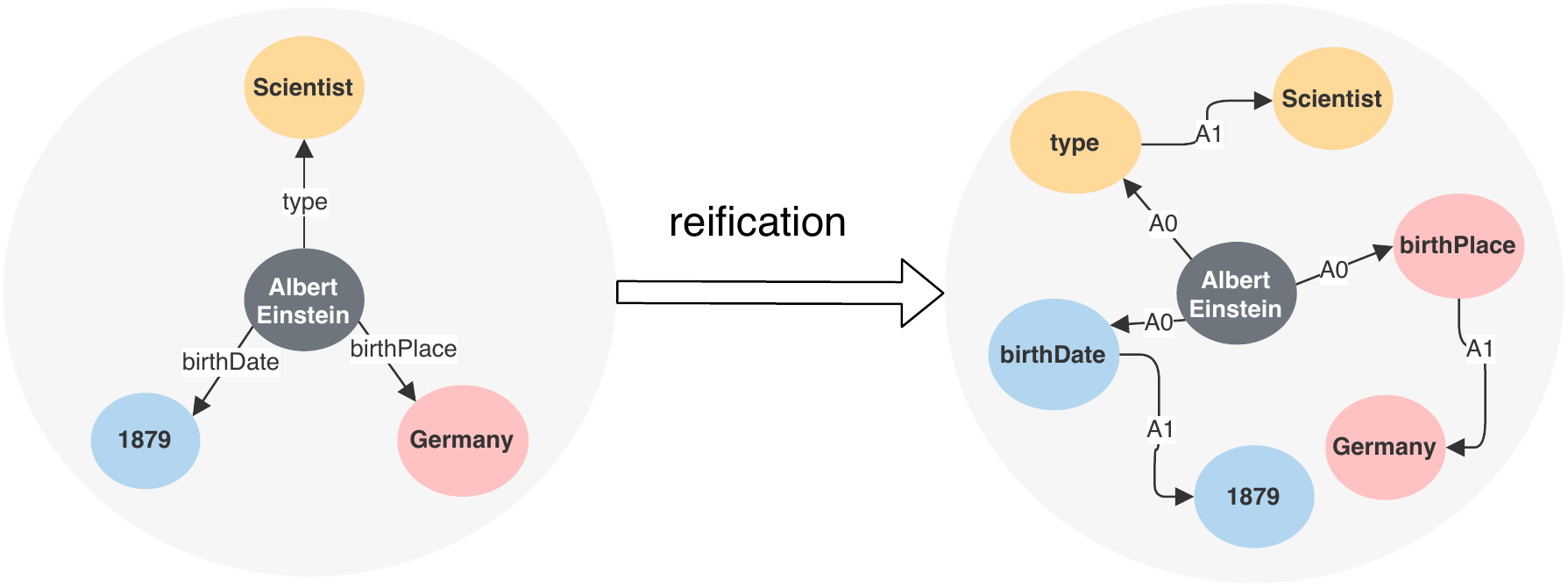}
\caption{Reification used on our example.}
\label{fig:reification}
\end{figure}

Therefore, we follow the reification strategy, which maps the relations to nodes in the \ac{KG} and creates new binary relations for each relation in the RDF triples. We rely on two binary relations, which model the relationship between the subject and predicate (A0) and predicate and object (A1) only. For example, \triple{Albert\_Einstein}{birthPlace}{Germany}
becomes \triple{Albert\_Einstein}{A0}{birthPlace} and  \triple{birthPlace}{A1}{Germany}. Apart from handling the parameter explosion problem, reification is useful in two ways. First, the encoder generates a hidden state for each relation in the input. Second, it allows for modeling an arbitrary number of edges (predicates) efficiently. Figure~\ref{fig:reification} illustrates the reification strategy for our example.

\subsubsection{Encoder.} 
\begin{figure}[htb]
\centering
\includegraphics[scale=0.7]{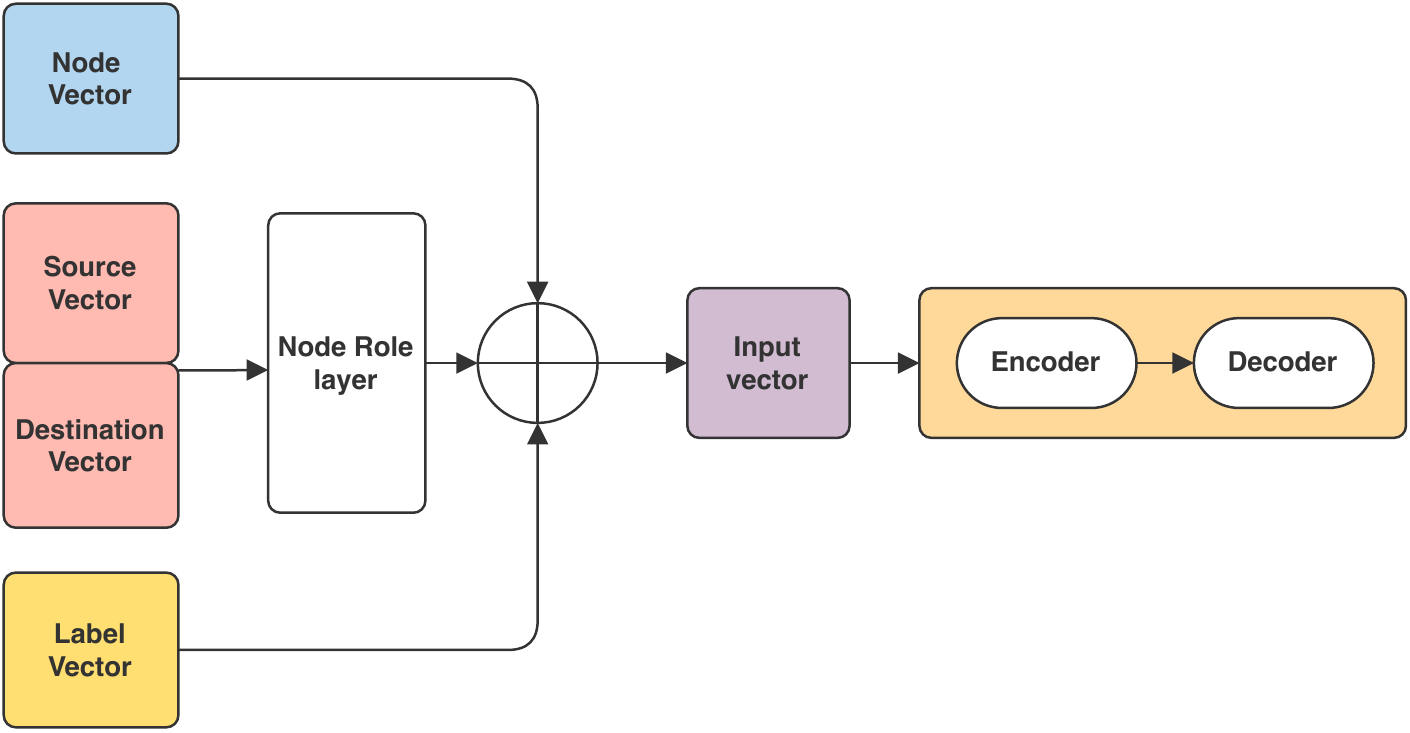}
\caption{An overview of a single forward pass in NABU.}
\label{fig:NABU_forward}
\end{figure}

Here, the reified graph is sent as input to the \ac{GAT} that applies a self-attention mechanism to compute the importance of each node in the graph. The \ac{GAT} encoder represents nodes in a high-dimensional vector space whilst taking into account the representations of their neighbors. Note that NABU follows the same strategy of recent literature on multilingual \ac{NMT} models in which a special token is used in the encoder to determine to what target language to translate~\cite{tan2019multilingual}. Figure~\ref{fig:NABU_forward} shows how a single forward step/pass works in NABU approach. 

In one forward pass of our model, we have four dense vectors as inputs, namely (i) the \emph{node vector}   $\Vec{H}=(\Vec{h_{1}}, \Vec{h_{2}},.., \Vec{h_{z}})$ with embeddings of all nodes in the graphs, (ii) the \emph{source vector}, $\Vec{S}=(\Vec{s_{1}}, \Vec{s_{2}},.., \Vec{s_{z}})$ with embeddings of source nodes in edges of the graph, (iii) the \emph{destination vector},  $\Vec{D}=(\Vec{d_{1}}, \Vec{d_{2}},.., \Vec{d_{z}})$ with embeddings of target nodes in edges of the graphs and (iv) the \emph{label vector}, $\Vec{L}=(\Vec{l_{1}}, \Vec{l_{2}},.., \Vec{l_{z}})$
with embedding labels. 
The source $\Vec{S}$ and destination $\Vec{D}$ vectors are concatenated and are passed through dense layer which encodes them into a vector of the same shape as the label vector. We call this new vector the \emph{edge vector}, $\Vec{E}$. We then add the edge vector ($\Vec{E}$), node vector ($\Vec{H}$) and label vector ($\Vec{L}$) to form the input vector to our encoder: 

$$\Vec{E} = f(\Vec{S}, \Vec{D}),\mbox{ and}$$
$$\Vec{H^{'}} = \parallel_{h \epsilon \eta} G(\Vec{H}+\Vec{L}+\Vec{E}),$$
where $\eta$ is the number of heads in the multi-head attention layer.

\subsubsection{Decoder} Our decoder follows the standard architecture of the Transformer decoder, which takes into account the intermediate representation generated by the encoder. The decoder gives a probability distribution over the target language's vocabulary. We also rely on an unsupervised tokenizer, which implements \ac{BPE}~\cite{sennrich2016b} and unigram language model~\cite{kudo2018subword} for handling  multilinguality and out-of-vocabulary words. Afterward, we apply a beam search for selecting the most likely word in the output sentence.

\section{Evaluation}

\subsection{Goals}
\label{sec:goals}

In our evaluation, we address the following research questions:

\begin{itemize}
\setlength\itemsep{0mm}

\item[Q1:] How does our multilingual approach compare with state-of-the-art results in English?

\item[Q2:] Is NABU able to generate bilingual text while modelling two languages from distinct families?

\item[Q3:] How accurate are the multilingual texts generated by NABU?

\end{itemize}

We designed our evaluation as follows: First, we measured the performance of NABU on English by using the WebNLG dataset and compared it with state-of-the-art approaches. Additionally, we evaluated NABU on two other languages---German and Russian. Second, we evaluated NABU on bilingual models ---English-German and English-Russian. Third, we combined all three languages in a multilingual setting and compared it with a multilingual Transformer baseline model. For measuring the quality of our approach, we used the automatic evaluation metrics {\sc BLEU}, {\sc METEOR}, and {\sc chrF++}~.

\subsection{Data}

The experiments presented in this work were conducted on the WebNLG corpus \cite{claire2017,claire2017b}, which consists of sets of RDF triples mapped to target texts. In comparison with other popular NLG benchmarks \cite{belz2011,novikova2017b,mille2018}, WebNLG is the most semantically varied corpus. Its English version contains 25,298 texts which describe 9,674 sets of up to 7 RDF triples in 15 domains: Astronaut, University, Monument, Building, Comics Character, Food, Airport, Sports Team, Written Work, City, Athlete, Artist, Means of Transportation, Celestial Body and Politician. Out of these domains, five ((Athlete, Artist, MeanOfTransportation, CelestialBody, Politician)) are exclusively present in the test set, being unseen during the training and validation processes. 

For German and Russian, we relied on the translated versions of WebNLG corpus~\cite{ferreira2018b,shimorina2019creating}. The German version comprises 20,370 texts describing 7,812 sets of up to 7 RDF triples in 15 domains. Additionally, the German datasets provide gold-standard representations for traditional pipeline steps, such as discourse ordering (i.e., the order in which the source triples are verbalized in the target text), text structuring (i.e., the organization of the triples into paragraph and sentences), lexicalization (i.e., verbalization of the predicates) and referring expression generation (i.e., verbalization of the entities). The Russian datasets contain 20,800 texts describing 5,185 sets of up to 7 RDF triples in 9 domains. Both were automatically created and manually analyzed. The English and Russian datasets abide by the criteria to gold standards as they were manually assessed by several native speakers. The German version can be regarded as a silver standard given that it did not go through the same process and contains some known errors.
 For the monolingual experiments, we relied on the standard WebNLG parts of train, dev, and test sets across all languages. Note that the German version does not contain a test set originally. Therefore we relied on a k-Fold Cross-Validation technique to create the test set. For the multilingual set of experiments, we concatenated all English, German and Russian datasets and shuffled their training sets randomly to facilitate an end-to-end training of the model. 

\subsection{Tasks}

We designed three tasks for carrying out our evaluation, (1) Monolingual, (2) Bilingual, (3) Multilingual. (1) In the monolingual task, we train our models to work in each language separately. Hence, we generate three models, one for English, one for German, and another for Russian. Each model receives RDF triples from its given DBPedia language version. For example, the German model receives triples from the German DBpedia. Afterward, we evaluate the models on each WebNLG language-specific dataset. (2) The bilingual task was divided into two sets; the first set, we train one English-German model. This model receives RDF triples from the English and German DBpedia versions as input and has to generate text in English and German, as output. For the second set, we trained one English-Russian model that receives RDF triples from the English and Russian DBpedia versions and generates text in English and Russian, respectively. (3) In the third task, we train one multilingual model which receives as input the triples from the English, German, and Russian DBpedia versions. This model has to output text in three languages, English, German, and Russian, respectively. The input relies on WebNLG triples containing resources from the English, German, and Russian DBpedia KGs, all entities are found across the three \acp{KG} via \texttt{sameAs} relations for the sake of completeness.

\subsection{Model settings} 

In this section, we describe the parameters and hyper-parameters used to train NABU models. We experimented with two encoder-decoder architectures for RDF verbalization. First, Transformer$_{baseline}$ which is an encoder-decoder model with a pure transformer architecture used to both encode triples into intermediate representation and decode it into tokens. Second, NABU$_{GAT-Trans}$, which comprises a \ac{GAT} encoder and Transformer as the decoder. 

For both models, we relied on the same settings. We used a Transformer 6-layer encoder-decoder model with an 8-headed multi-head attention mechanism~\cite{vaswani2017}. The training used a batch size of 32 and Adam optimizer with an initial maximum learning rate of 0.001. We set a source and target word embedding's size of 256, and hidden layers to size 256, dropout = 0.3 (naive). We used a vocabulary of 32000 words for the word based models and a beam size of 5. All our vocabularies were trained using the sentencepiece library.\footnote{\url{https://github.com/google/sentencepiece}} In addition, we used a copy mechanism for investigating the \ac{OOV} words issue. This mechanism first tries to substitute the \ac{OOV} words with target words that have the highest attention weight according to their source words~\cite{LuongAttention2015}. If the words are not found, it copies the source words to the position of the not-found target word~\cite{gu2016incorporating}.  Note that we added an extra language token at the beginning of our input sentences for the Transformer model, and a language node to the input graph in our GAT model for performing the bilingual and multilingual experiments. This technique of adding a special language token is in line with~\cite{tan2019multilingual}.


\subsection{Evaluation Metrics}

We used three automatic \ac{MT} standard metrics to ensure consistent and clear evaluation of the common evaluation datasets of the WebNLG challenge. {\sc BLEU}~\cite{papineni2002bleu} uses a modified precision metric for comparing the \ac{MT} output with the reference (human) translation. The precision is calculated by measuring the n-gram similarity (n=1,..4) at the word level. BLEU also applies a brevity penalty by comparing the length of the \ac{MT} output with the reference translation. { \sc METEOR}~\cite{banerjee2005meteor} was mainly introduced to overcome the semantic weakness of {\sc BLEU}. To this end, METEOR considers stemming and paraphrasing along with exact standard word (or phrase) matching. The synonymy overlap through a shared WordNet synset of the words. Along with exact standard word (or phrase) matching, it has additional features, i.e., stemming and paraphrasing. {\sc chrF++}~\cite{popovic2017chrf++} exploits the use of character n-gram precision and recall (F-score) for automatic evaluation of \ac{MT} outputs. chrF++ has shown a good correlation with human rankings of different \ac{MT} outputs and is simple and does not require any additional information. Additionally, chrF++ is language- and tokenization-independent. 

\subsection{Results}

\paragraph{\textbf{Monolingual.}} Our experiments report that NABU consistently outperforms state-of-the-art models on English data. Table~\ref{tab:english_results} shows that NABU achieved a BLEU score of 66.21, which is 28.15\% higher than the previous state-of-the-art Transformer model~\cite{ferreira2019neural}. We decided to run our experiments on all WebNLG categories to elucidate the strengths and limitations of NABU. According to~\cite{ferreira2019neural}, the main drawback in current \ac{NN} models is the incapability of generating text for unseen entities and that the experiments should be on all categories. NABU, in turn, shows that it is capable of predicting correctly both seen and unseen entities and their relations. In addition, NABU shows an improvement in METEOR up to +2 points. We report NABU's chrF++ as our intention is to follow recent literature which has adopted this metric due to its good correlation with human results. We can now answer [Q1] as follows: NABU surpasses state-of-the-art results on WebNLG in English.   

\begin{table}[!htb]
\centering
\setlength\tabcolsep{4pt}
\caption{Results on WebNLG English test set with all categories (seen and unseen), comparison with the state-of-the-art approaches}
\begin{tabular}{@{}lccc@{}}
\toprule
\textbf{Model} & \textbf{BLEU} & \textbf{METEOR} & \textbf{chrF}++ \\
\toprule
UPF-FORGe & 38.65 & 39.00 & - \\
Melbourne & 45.13 & 37.00 & - \\
Moryossef et al., 2019) & 47.40 & 39.00 & - \\
Castro et al. (2019) & 51.68 & 32.00 & - \\
\bottomrule
NABU$_{GAT-Trans}$ & \textbf{66.21} & \textbf{41.11} & \textbf{71.98} \\
\bottomrule
\end{tabular}
\label{tab:english_results}
\end{table}

 Table~\ref{monolingua_results_german_russian} shows that NABU outperforms the transformer baseline on German and Russian. It is important to note that our Transformer baseline, Transformer$_{baseline}$, already outperforms the previous state-of-the-art approaches on English. The difference between our Transformer$_{baseline}$ and the Transformer presented by \cite{ferreira2019neural} is that we rely on \ac{BPE} and character-level tokenizer on the decoder side. Our results suggest that we can refrain from running the related work (see Table~\ref{tab:english_results}) on the German and Russian datasets, especially as they were designed and tested to work on English, thus there is currently no baseline for German and Russian. With these results, NABU demonstrates its language agnosticism and presents improvements in German and Russian over the baseline.

\begin{table}[!htb]
\centering
\setlength\tabcolsep{4pt}
\caption{Monolingual Results on WebNLG language testsets}
\begin{tabular}{@{}llccc@{}}
\toprule
\textbf{Models} & \textbf{Language} & \textbf{BLEU} & \textbf{METEOR} & \textbf{chrF++} \\
\toprule
\multicolumn{5}{c}{ \textbf{Monolingual} } \\
\toprule
\multirow{3}{*}{ Transformer$_{baseline}$ } & ENG & 54.96 & 38.43 & 69.11 \\
& GER & 50.07 & 34.51 & 63.48 \\
& RUS & 46.42 & 27.74 & 56.80 \\
\midrule
\multirow{3}{*}{ NABU$_{GAT-Trans}$ } & ENG & \textbf{66.21} & \textbf{41.47} & \textbf{71.98} \\
& GER & \textbf{53.08} & \textbf{37.42} & \textbf{64.57} \\
& RUS & \textbf{46.86} & \textbf{28.84}& \textbf{58.37}\\
\bottomrule
\end{tabular}
\label{monolingua_results_german_russian}
\end{table}

\paragraph{\textbf{Bilingual.}} Table~\ref{bilingual_results} presents the results of NABU$_{GAT-Trans}$ on two bilingual models. The results show that NABU on English-German outperformed the Transformer$_{baseline}$ on all metrics. On English-Russian, NABU$_{GAT-Trans}$ presented worse results on BLEU and METEOR than Transformer$_{baseline}$. However, NABU$_{GAT-Trans}$ showed superior results on chrF++ which is the metric that best correlates with human results. On the one hand, we analyzed that the English-German model leveraged both languages properly due to their vocabulary overlap. German and English share a word vocabulary of 33\%, thus training both languages with NABU$_{GAT-Trans}$, which employs a graph representation on the encoder side and a character level on decoder could actually model both languages correctly and generate coherent text. On the other hand, English-Russian presented inconsistent results because both languages are significantly different, and they do not share any vocabulary. We reckoned these conflicting scores are due to the language family of both languages. Looking manually at the results, we concluded that encoding distinct language families requires additional features, and we, therefore, plan to investigate this phenomenon in the future. The results presented herein answer our second research question, [Q2], by showing that NABU is capable of modeling languages from distinct families in a bilingual approach, but a deeper investigation is required.

\begin{table}[!htb]
\centering
\setlength\tabcolsep{4pt}
\caption{Bilingual Results on WebNLG language test sets}
\begin{tabular}{@{}llccc@{}}
\toprule
\textbf{Models} & \textbf{Language} & \textbf{BLEU} & \textbf{METEOR} & \textbf{chrF++} \\
\toprule
\multicolumn{5}{c}{ \textbf{Bilingual} } \\
\toprule
Transformer$_{baseline}$ & ENG-GER & 58.30 & 36.46 & 66.72 \\
NABU$_{GAT-Trans}$ & ENG-GER & \textbf{61.99} & \textbf{39.51} & \textbf{69.68} \\
\midrule
Transformer$_{baseline}$ & ENG-RUS & \textbf{55.30} & \textbf{37.90} & 61.63 \\
NABU$_{GAT-Trans}$ & ENG-RUS & 49.15 & 33.41 & \textbf{64.00} \\
\bottomrule
\end{tabular}
\label{bilingual_results}
\end{table}

\paragraph{\textbf{Multilingual.}} Table~\ref{multilingual_results} shows that NABU$_{GAT-Trans}$ performed better than Transformer\-$_{baseline}$ by presenting consistent improvement of +2 BLEU, METEOR, and chrF++. This result exhibits that NABU can effectively generate multilingual text, thus answering our third research question, [Q3]. Comparing the multilingual results of NABU with its bilingual results on English-Russian, we concluded that the characteristics of the German language, namely its three gender types, contributed to the better alignment of the languages in the decoder side of multilingual NABU model. Russian also contains three genders as German; therefore, NABU made use of it as features for generating coherent texts. We also noticed that the English texts generated by the multilingual NABU model are comparable to those of the English state-of-the-art models. NABU's multilingual model is also better than the previous English state-of-the-art by 4 BLEU and presents comparable results on METEOR. This result also reaffirms the capability of NABU for achieving English state-of-the-art results and contributes to our first research question, [Q1].      

\begin{table}[!htb]
\centering
\setlength\tabcolsep{4pt}
\caption{Multilingual Results on WebNLG language testsets}
\begin{tabular}{@{}llccc@{}}
\toprule
\textbf{Models} & \textbf{Language} & \textbf{BLEU} & \textbf{METEOR} & \textbf{chrF++} \\
\toprule
\multicolumn{5}{c}{ \textbf{Multilingual} } \\
\toprule
Transformer$_{baseline}$ & ENG-GER-RUS & 53.39 & 36.86 & 60.72 \\
NABU$_{GAT-Trans}$ & ENG-GER-RUS & \textbf{56.04} & \textbf{38.34} & \textbf{62.04} \\
\bottomrule
\end{tabular}
\label{multilingual_results}
\end{table}

\paragraph{\textbf{Time-Performance.}} All models were trained on NVIDIA Tesla P100. Both NABU\-$_{GAT-Trans}$ and Transformer$_{baseline}$ models took the same amount of time since they contain the same number of weights. Therefore, the monolingual models took 6 hours to be trained, while the multilingual models took 8 hours on average. This difference of 2 hours lies in the size of the multilingual training dataset, which contains all English, German, and Russian training sets.

\subsection{Error Analysis and Discussion}
\label{sec:error}

In this section, we report some of the errors found in NABU's output while carrying out a human evaluation. First, we analyzed the discrepancy between BLEU, METEOR, and chrF++: NABU outperformed the previous state-of-the-art approach for English by roughly 15 BLEU, while the difference in METEOR is considerable smaller.  
Our analysis shows that some entities contained typos and were not generated correctly by NABU. In addition, we found a low variance in the generated synonyms. BLEU ignores these aspects while METEOR penalizes based on them, thus explaining the discrepancy between the scores. 

Additionally, we noticed some wrong verbalization of similar predicates (edges) that were responsible for decreasing NABU scores across all languages. For example, NABU was sometimes not able to generate text correctly in the Artist domain. The problem lies in the triples which contain both \texttt{dbo:artist} or \texttt{dbo:producer} relations as predicates. Both predicates are often verbalized to ``artist''. This happens because the predicates share the same domain and range and therefore have a similar vector representation in the embeddings. We plan to address this issue in future work by using a more appropriate embedding model. 

We also analyzed the multilingual texts generated by NABU$_{GAT-Trans}$ and Transfor\-mer$_{baseline}$. We noticed that the NABU$_{GAT-Trans}$ performed better at structuring the RDF graph as input and verbalizing a structured set of RDF triples, whereas Transformer$_{baseline}$ presented better results than NABU$_{GAT-Trans}$   
at ordering (also known as Discourse Ordering step) the triples for a better verbalization. The advantage of Transformer$_{baseline}$ over NABU\-$_{GAT-Trans}$ in Discourse Ordering seems to be related to the linearized form of its input, which explicitly represents in what order the triples have to be verbalized. Additionally, our reification strategy affected the Discourse Ordering, we noticed it by analyzing the generated text from an input with two equal predicates for different subjects. For example, ``Albert\_Einstein dbo:birthPlace Germany'' and ``Michael\_Jackson dbo:birthPlace USA''. NABU$_{GAT-Trans}$ verbalized this two triples as ``Albert Einstein was born in the United States of America and Michael Jackson was born in Germany''. This problem occurs because NABU can not identify the subjects of each predicate correctly as they are identical in the encoder side. We plan to address this drawback by investigating new approaches for the structuring and ordering steps.




Another interesting insight is related to the inflections of words in German, similar to \cite{ferreira2018b}. The possessive was often a source of errors when verbalizing into German. The translation ``Elliot See 's Besatzung war ein Testpilot.'' is not perfect as the apostrophe ('s) is placed wrongly. However, this problem did not happen when generating the sentence, ``Bill Oddies Tochter ist Kate Hardie'', where the possessive of ``Oddie'' is built correctly. Similar insights can be derived pertaining to the preposition ``von'' (en: of). For example, the entity \texttt{Texas\_University} was wrongly verbalized as ``Universit\"at von Texas'' instead of the correct form ``Universit\"at Texas''. The possessive and related constructions are well-known challenges in \ac{MT} from English to German. 
Therefore, we plan to explore this phenomenon in future research deeply.

On the Russian results, we observed that the main challenge was related to the verbalization of unseen entities. In NABU$_{GAT-Trans}$, some entities were copied from their source sentences due to the use of the copy mechanism in NABU. For example, the entity ``Visvesvaraya\_Technological\_University'' was generated as ``Visvesvaraya Technical University'' in the English form instead of being verbalized in the Russian language. Additionally, we perceived that NABU$_{GAT-Trans}$ displayed problems similar to those reported in~\cite{shimorina2019creating} for generating Entities. However, these problems were mostly detected in the unseen category. Our current hypothesis is that the generation of unseen entities in Russian is more challenging than German and English due to the Cyrillic alphabet.

\section{Conclusion}

We presented a multilingual RDF verbalizer which relies on graph attention \ac{NN} along with a reification strategy. 
Our experiments suggest that our approach, named NABU, outperforms state-of-the-art approaches in English. Additionally, NABU presented consistent results across the languages used in our evaluation. NABU is language-agnostic, which means it can be ported easily to languages other than those considered in this paper. 
To the best of our knowledge, we are the first approach to exploit and achieve the multilinguality successfully in the RDF-to-text task. As future work, we aim to exploit other graph-based neural architecture and other reification approaches for improving NABU's performance. Additionally, we plan to investigate how to deal with the similarity of relations by combining language models and new evaluation metrics~\cite{bleurt}. Moreover, we plan to investigate our methodology in the context of low-resource scenarios as well as on different \acp{KG}~\cite{kaffee2018mind,rdf2pt_lrec_2018}.

\section*{Acknowledgments} 	

Research funded by the German Federal Ministry of Economics and Technology (BMWI) in the project RAKI (no. 01MD19012D) and by the H2020 KnowGraphs (GA no. 860801). This work also has been supported by the German Federal Ministry of Education and Research (BMBF) within the project DAIKIRI under the grant no 01IS19085B as well as by the German Federal Ministry for Economic Affairs and Energy (BMWi) within the project SPEAKER under the grant no 01MK20011U. Finally, we also would like to thank the funding provided by the Coordination for the Improvement of Higher Education Personnel (CAPES) from Brazil under the grant 88887.367980/2019-00.


\footnotesize
\bibliographystyle{plain}
\bibliography{bibliography}

\end{document}